\documentclass{article}
\usepackage{spconf,amsmath,graphicx,hyperref,amssymb,amsfonts}
\usepackage{algorithmic}
\usepackage{graphicx}
\usepackage{textcomp}
\usepackage{xcolor}
\usepackage{tabularx}
\usepackage{booktabs}
\usepackage{multirow}
\usepackage{graphicx} 
\usepackage{xcolor}
\usepackage{bm}
\usepackage{hyperref} 

\title{SPLF-SAM: Self-Prompting Segment Anything Model for Light Field Salient Object Detection}
\name{Qiyao Xu$^{\dagger}$\quad Qiming Wu$^{\dagger}$\quad Xiaowei Li$^{*}$\thanks{$^\dagger$  These authors contribute equally to this work. $^{*}$Xiaowei Li is the corresponding author with E-mail: xwli@scu.edu.cn. Thos work is supported by the National Natural Science Foundation of China (No. 62275177).}}

\address{College of Electronic Engineering and Information, Sichuan University, Chengdu, China}
\begin{document}
%
\maketitle
\begin{abstract}
Segment Anything Model (SAM) has demonstrated remarkable capabilities in solving light field salient object detection (LF SOD). However, most existing models tend to neglect the extraction of prompt information under this task. Meanwhile, traditional models ignore the analysis of frequency-domain information, which leads to small objects being overwhelmed by noise. In this paper, we put forward a novel model called self-prompting light field segment anything model (SPLF-SAM), equipped with unified multi-scale feature embedding block (UMFEB) and a multi-scale adaptive filtering adapter (MAFA). UMFEB is capable of identifying multiple objects of varying sizes, while MAFA, by learning frequency features, effectively prevents small objects from being overwhelmed by noise. Extensive experiments have demonstrated the superiority of our method over ten state-of-the-art (SOTA) LF SOD methods. Our code will be available at \url{https://github.com/XucherCH/splfsam}.
\end{abstract}
\section{Introduction}
\label{sec:intro}

Salient Object Detection (SOD)\cite{10446685} is a task rooted in the visual attention mechanism, aiming to identify objects or regions that attract the most attention within an RGB image. As one of the most important downstream tasks in computer vision, SOD is widely applied in fields such as Segmentation and Object Detection. However, the vast majority of existing models are designed for conventional two-dimensional images. To address more complex real-world detection tasks, it is imperative to utilize light field all-focus images which preserve essential 3D information such as depth cues compared to standard 2D images. This task is widely known as light field salient object detection (LF SOD)\cite{JIANG2025117}.

Unlike traditional images capturing only spatial data, light field (LF) cameras simultaneously acquire both spatial and angular information. This makes it possible to process images with more complex backgrounds. With the development of deep learning, LF SOD have seen significant progress. Researchers have proposed many models based on backbone architectures such as convolutional neural networks (CNNs), Transformers, and Mamba\cite{zhu2024visionmambaefficientvisual}, which have demonstrated strong performance on existing datasets. In further studies, multi-scale enhanced feature fusion\cite{huang2024saliency} has also provided new insights for the development of LF SOD.
During the past several years, driven by advancements in large vision models like Segment Anything Model (SAM), LF SOD has also achieved remarkable breakthrough. Gao et al.\cite{gao2024multiscaledetailenhancedsegmentmodel} introduce a prompt-free SAM network (MDSAM). However, in the absence of prompt and multi-level decoders, they do not fully utilize the semantic information from the encoder. Additionally, the lack of frequency component analysis also leads to some targets being drowned out by noise. Building from previous SAM researches, Zhang et al.\cite{rs17020342} put forward a self-prompting SAM  for its decoders. But the low resolution of the first prompt map causes smaller targets to be obscured by noise. His method fails to adequately integrate multi-level prompt information nor frequency information as well.

To overcome these limitations, we are the first to introduce a novel approach called self-prompting light field segment anything model(SPLF-SAM). Targeting the characteristics of LF SOD, we present a multi-scale adaptive filtering adapter to achieve efficient fine-tuning. Specifically, we propose a unified multi-scale feature embedding block (UMFEB). It helps feature such as prompt and image embeddings fusion globally without any changes to this block. Meanwhile, inspired by SAM2, we introduce a prompt bank and a decoder with convolutional gating block to store and analysis all prompt information. 

In summary, our main contributions include:

\textbf{(1)} We create a multi-scale adaptive filtering adapter (MAFA) for efficient fine-tuning to efficiently learn features, especially small features.

\textbf{(2)} We propose a unified multi-scale feature embedding block (UMFEB) to separate targets from background and noise without any changes to this block.

\textbf{(3)} The proposed SPLF-SAM, equipped with Prompt Bank and multi-level decoders, achieves state-of-the-art (SOTA) performance on multiple datasets, demonstrating the superiority of our approach.
\begin{figure*}[!htbp]
    \centering
    \includegraphics[width=0.7\linewidth]{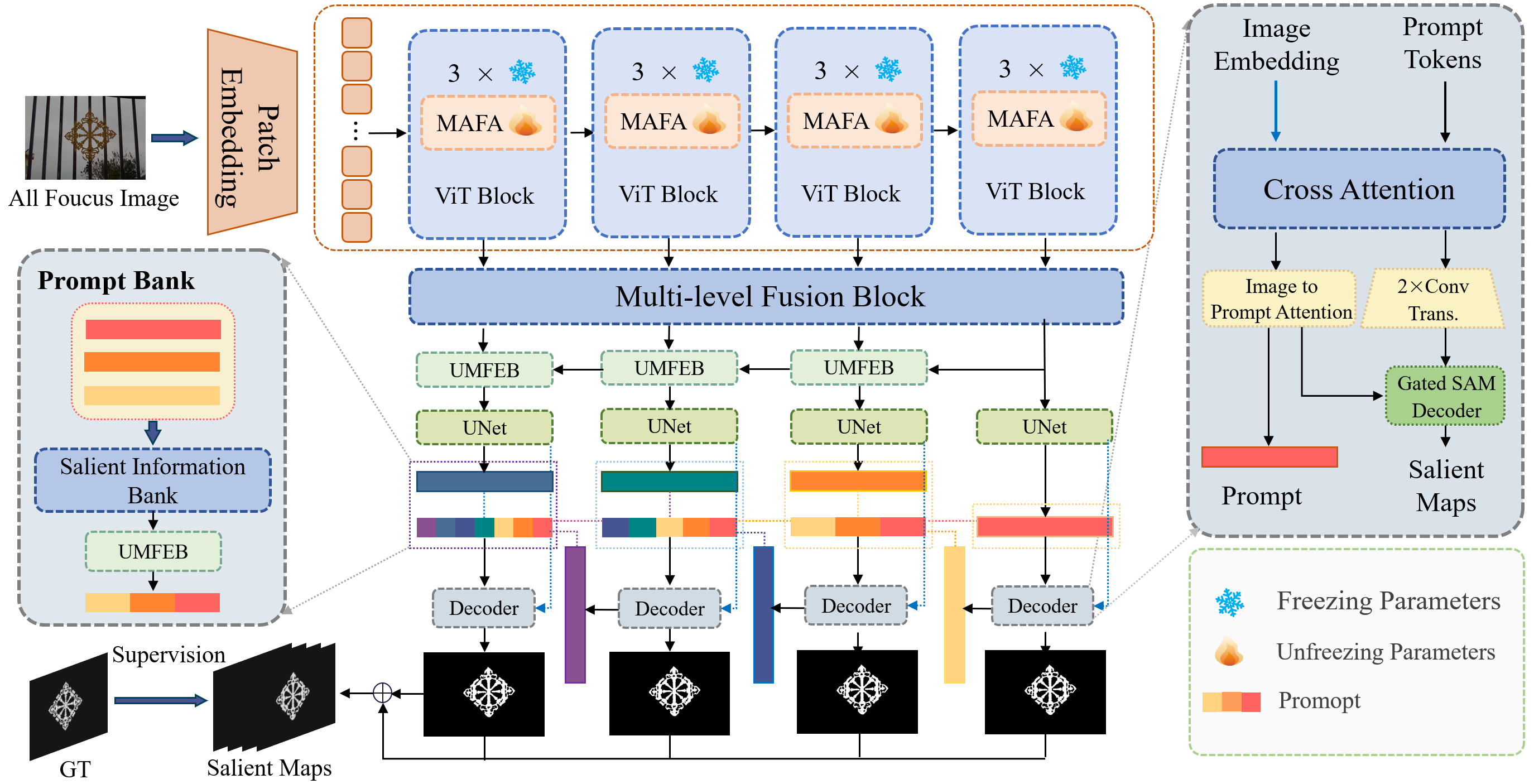}
    \caption{The overall flowchart of the self-prompting light field segment anything model (SPLF-SAM).}
    \label{framework}
    \vspace{-0.5cm}
    \label{fig:1}
\end{figure*}

\section{PROPOSED METHOD}
\label{sec:format}

\subsection{Overview}

The overall of our proposed SPLFSAM is illustrated in Fig. \ref{fig:1}, in which all encoder weights except proposed adapters are frozen and excluded from training. Consistent with the original SAM implementation, we only extract encoder outputs from the 2nd, 5th, 8th, and 11th layers, denoted as $\bm{R}_{1}$, $\bm{R}_{2}$, $\bm{R}_{3}$, $\bm{R}_{4}$. After extracting features from the encoders, the deepest feature $\bm{R}_{4}$. is first reshaped and undergoes U-NET block before being stored in the prompt bank. The information from the prompt bank is then fused with the features and fed into the decoder. The decoder generates both saliency map predictions and new prompt information. We merge the new prompts with existing ones and update the prompt bank. Meanwhile, adjacent high-level and low-level features are fused to ensure information propagation. This process iterates for $\bm{R}_{i}$ sequentially. All fusion operations are implemented through our proposed UMFEB module. Please note that prompt information from the  $\bm{R}_{1}$ decoder output is discarded. Details of the module are described below.
\subsection{Multi-scale Adaptive Filtering Adapter}
\begin{figure}[htbp]
\centering
\includegraphics[width=0.9\columnwidth]{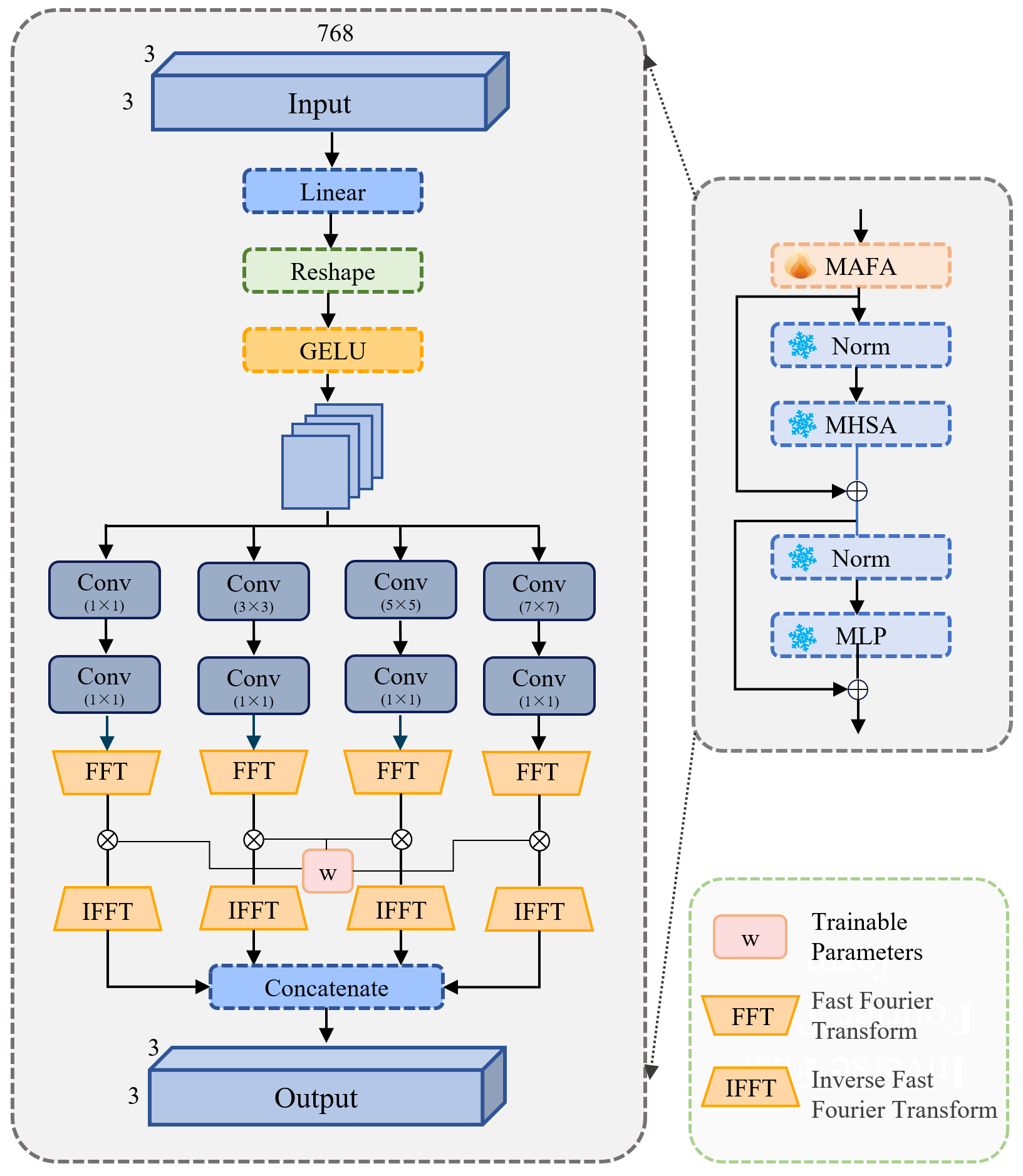}
\caption{The illustration of multi-scale adaptive filtering adapter (MAFA) .}
\label{fig:2}
\end{figure}

Traditional adapter-based fine-tuning commonly employs a downsampling–upsampling scheme. This process tends to introduce noise, especially when dealing with small targets, causing detection failure. To address this, we develop a multi-scale adaptive filtering adapter (MAFA) in Fig. \ref{fig:2}. Let $\mathbf{F}$ denote input feature from encoder. After passing through MLP, the features will subsequently pass through convolutions with different kernel sizes, respectively. For each feature map, we partition the image into patches of size 8 following the ViT methodology. Each patch undergoes Fourier transform. Concurrently, we introduce learnable kernels for filtering, followed by inverse Fourier transform to reconstruct the original image. The fused feature and filtering process of the subnetwork can be generated as follows: 

\begin{equation}
\begin{split}
    &\bm{F}_{i}^{\mathit{Frequency}} = \Phi(\operatorname{GELU}(\operatorname{Conv}_{i}(\bm{F}_{i}))), \\
    &\bm{F}_{i}^{\mathit{Filtered}} = \Phi^{-1}(\bm{W}_{i}^{\mathit{Learnable}} \cdot \bm{F}_{i}^{\mathit{Frequency}}),
\end{split}
\end{equation}
where $\Phi$($\cdot$) denotes the process of FFT. After this, the filtered results are concatenated and then the channels are adjusted using convolution. 

\subsection{Unified Multi-scale
Feature Embedding Block}

\begin{figure}[htbp]
\centering
\includegraphics[width=0.9\columnwidth]{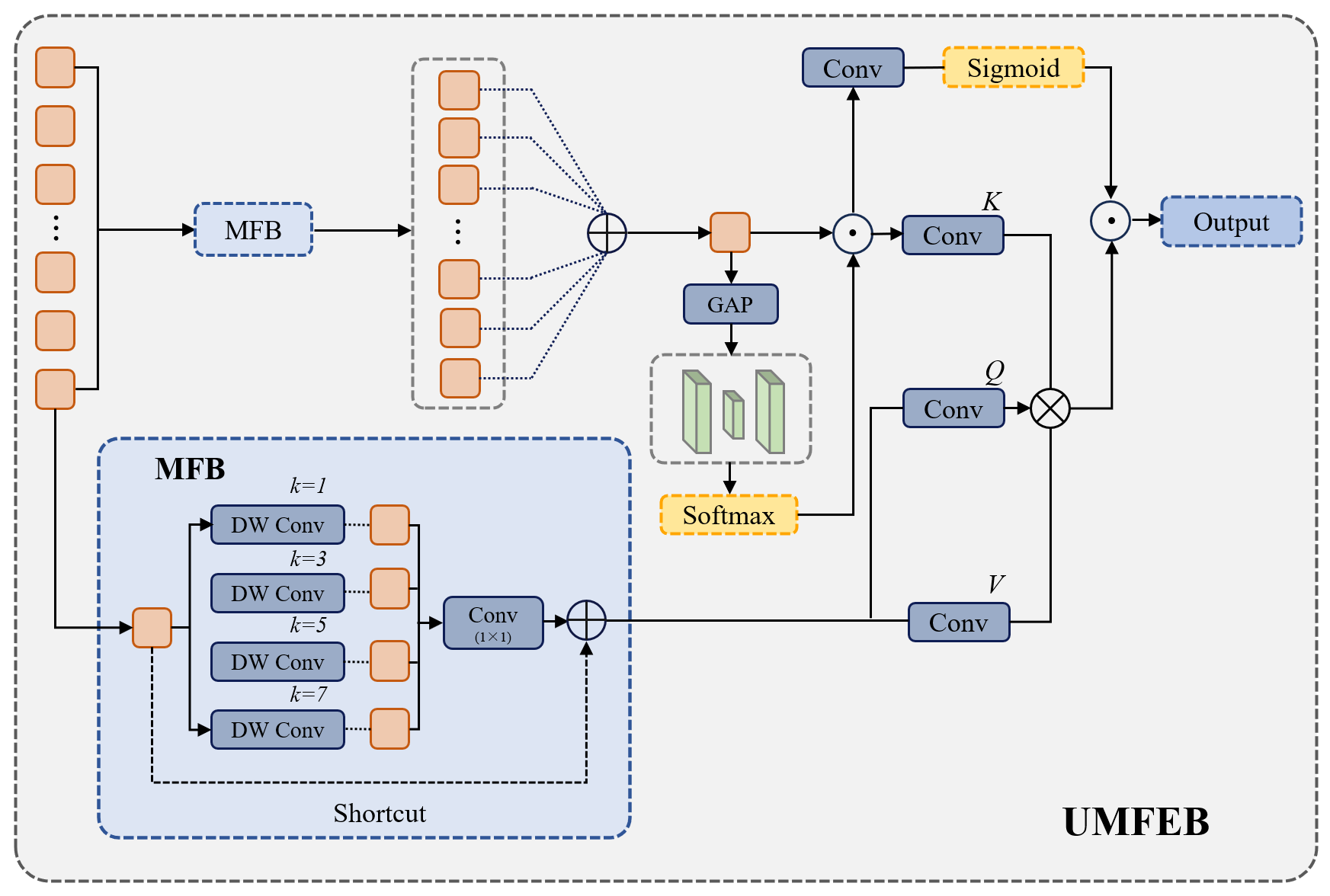}
\caption{The illustration of unified multi-scale feature embedding block (UMFEB).}
\label{fig:3}
\end{figure}

The current approach of concatenation followed by convolution offers straightforward implementation. However it suffers from limited receptive fields which leads to insufficient feature fusion. To address these issues, we develope a unified multi-scale feature embedding block (UMFEB). The detailed structure of UMFEB can be observed in Fig. \ref{fig:3}. Let $\mathbf{F}_{i}$ denote features for fusion, where $\mathbf{F}_{1}$ represents the most recently input feature. We first refine $\mathbf{F}_{1}$ and all $\mathbf{F}_{i}$ which have been concatenated with multiscale fusion block (MFB). Those processes are shown as follows:

\begin{equation}
\begin{split}
    &\bm{F}^{\mathit{MFB}}_{\mathit{1}} = \operatorname{MFB(\bm{F}_{1})}, \\
    &\bm{F}^{\mathit{MFB}}_{i} = \operatorname{MFB(\mathit{Concat}(\bm{F}_{\mathit{i}}))}.
\end{split}
\end{equation}
After those operations, $\bm{F}^{\mathit{MFB}}_{i}$ is fused with channel attention. Then $\bm{F}^{\mathit{MFB}}_{\mathit{1}}$ will serve as value ($\bm{V}$) and query ($\bm{Q}$), while $\bm{F}^{\mathit{MFB}}_{i}$ serves as key ($\bm{K}$). Meanwhile, $\bm{F}^{\mathit{MFB}}_{i}$ is also passed through a convolution operation and normalized using a sigmoid function to obtain an attention map. Finally, the output can be expressed as:

\begin{equation}
    \bm{F}^{\mathit{Output}}= \Omega(\operatorname{\operatorname{{Conv}_{1}(\bm{F}^{\mathit{MFB}}_{\mathit{i}})}})\cdot(\bm{Q}\bm{K})\bm{V}\bm{+}\bm{F}_{1}^{\mathit{Shortcut}},
\end{equation}
where $\Omega$ donotes Sigmoid function.

The Multiscale Fusion Block (MFB) enhances feature representation by capturing spatial context at multiple receptive fields. Given an input feature map \( \bm{F} \), MFB first applies a \(1 \times 1\) convolution, followed by four parallel depthwise convolutions with kernel sizes \(1 \times 1\), \(3 \times 3\), \(5 \times 5\), and \(7 \times 7\). The outputs are concatenated and fused via another \(1 \times 1\) convolution, then added to the input through a residual connection:
\begin{equation}
\bm{F}^{\mathit{MFB}} = \bm{F}^{\mathit{Shortcut}} + \operatorname{Conv}_{1 }\left(\operatorname{Cat}(\bm{F}_1, \bm{F}_3, \bm{F}_5, \bm{F}_7)\right),
\end{equation}
where \( \bm{F}_k \) denotes the output of the \(k \times k\) depthwise convolution.

\section{EXPERIMENTAL SETUP}
\label{sec:pagestyle}

\subsection{Implementation details}
\begin{table*}[t]
\centering
\small
\setlength{\tabcolsep}{2.3pt} 
\caption{Performance Comparison of Light Field Salient Object Detection Methods} 
\scalebox{0.92}{
\begin{tabular}{@{}l *{16}{c} @{}}
\toprule
 & \multicolumn{4}{c}{PKU-LF\cite{gao2023thorough}} & \multicolumn{4}{c}{DUT-LF\cite{wang2019deep}} & \multicolumn{4}{c}{HFUT\cite{zhang2017saliency}} & \multicolumn{4}{c}{Lytro Illum\cite{zhang2020light}} \\
\cmidrule(lr){2-5} \cmidrule(lr){6-9} \cmidrule(lr){10-13} \cmidrule(lr){14-17}
 & $S_\alpha\uparrow$ & $\quad F_\beta\uparrow$ & $\quad E_\phi\uparrow$ & $\quad M\downarrow$ & $S_\alpha\uparrow$ & $\quad F_\beta\uparrow$ & $\quad E_\phi\uparrow$ & $\quad M\downarrow$ & $S_\alpha\uparrow$ & $\quad F_\beta\uparrow$ & $\quad E_\phi\uparrow$ & $\quad M\downarrow$ & $S_\alpha\uparrow$ & $\quad F_\beta\uparrow$ & $\quad E_\phi\uparrow$ & $\quad M\downarrow$ \\
\midrule
Ours  & \color{red}0.943 & \color{red}0.955 & \color{red}0.981 & \color{red}0.018 & \color{blue}0.947 & \color{red}0.974 & \color{red}0.980 & \color{red}0.015 & \color{red}0.879 & \color{red}0.885 & \color{red}0.959 & \color{red}0.037 & \color{red}0.926 & \color{red}0.945 & \color{red}0.976 & \color{red}0.022  \\
GMERNet\cite{li2024gated}    & \color{blue}0.924 & \color{blue}0.918 & \color{blue}0.958 & \color{blue}0.026& \color{red}0.954 & \color{blue}0.957 & \color{blue}0.973 & \color{blue}0.018 & \color{blue}0.878 & \color{blue}0.852 & \color{blue}0.920 & \color{blue}0.046 & \color{blue}0.907 & \color{blue}0.896 & \color{blue}0.951 & \color{blue}0.030  \\
STSA\cite{gao2023thorough}       & 0.887 & 0.878 & 0.930 & 0.035& 0.928 & 0.941 & 0.965 & 0.027 & 0.834 & 0.810 & 0.884 & 0.057 & 0.890 & 0.888 & 0.936 & 0.037  \\
MEANet\cite{jiang2022meanet}     & 0.912 & 0.900 & 0.941 & 0.031 & 0.914 & 0.921 & 0.945 & 0.039 & 0.728 & 0.647 & 0.784 & 0.103 & 0.884 & 0.866 & 0.924 & 0.041  \\
JLDCF\cite{fu2021siamese}     & 0.854 & 0.811 & 0.889 & 0.049  & 0.877 & 0.878 & 0.925 & 0.058 & 0.789 & 0.727 & 0.844 & 0.075 & 0.881 & 0.868 & 0.926 & 0.044  \\
UCNet\cite{zhang2021uncertainty}     & 0.792 & 0.736 & 0.852 & 0.070 & 0.831 & 0.816 & 0.876 & 0.081 & 0.748 & 0.677 & 0.804 & 0.090 & 0.852 & 0.827 & 0.899 & 0.053  \\
D3Net\cite{fan2020rethinking}      & 0.802 & 0.742 & 0.858 & 0.067 & 0.822 & 0.797 & 0.860 & 0.083 & 0.749 & 0.671 & 0.797 & 0.091 & 0.860 & 0.836 & 0.905 & 0.055  \\
SANet\cite{zhang2021learning}     & 0.905 & 0.896 & 0.944 & 0.033 & 0.855 & 0.858 & 0.906 & 0.066 & 0.789 & 0.742 & 0.841 & 0.084 & 0.813 & 0.776 & 0.867 & 0.061  \\
DLGLRG\cite{liu2021light}    & 0.849 & 0.808 & 0.889 & 0.053 & 0.929 & 0.938 & 0.961 & 0.030 & 0.771 & 0.702 & 0.843 & 0.068 & 0.872 & 0.845 & 0.918 & 0.045  \\
BBS\cite{fan2020bbs}       & 0.847 & 0.802 & 0.882 & 0.056 & 0.865 & 0.852 & 0.900 & 0.066 & 0.751 & 0.676 & 0.801 & 0.089 & 0.876 & 0.848 & 0.909 & 0.047  \\
S2MA\cite{liu2020learning}      & 0.765 & 0.683 & 0.811 & 0.100  & 0.787 & 0.754 & 0.839 & 0.102 & 0.729 & 0.650 & 0.777 & 0.112 & 0.856 & 0.832 & 0.903 & 0.060  \\
ERNet\cite{piao2020exploit}     & 0.826 & 0.776 & 0.883 & 0.059 & 0.900 & 0.908 & 0.949 & 0.040 & 0.778 & 0.722 & 0.841 & 0.082 & 0.844 & 0.827 & 0.910 & 0.057  \\
MoLF\cite{wang2019deep}      & 0.809 & 0.776 & 0.883 & 0.066 & 0.887 & 0.903 & 0.939 & 0.052 & 0.742 & 0.662 & 0.811 & 0.095 & 0.834 & 0.820 & 0.908 & 0.065  \\
\bottomrule
\end{tabular}}
\label{tab:1}
\end{table*}

\noindent\textbf{Loss.} In this paper we use Binary Cross Entropy (BCE) with Logits Loss to supervise the four saliency predictions. The final loss function is defined as:

\begin{equation}
\mathcal{L}^{\text{Total}} = \sum_{i=1}^{4} \mathcal{L}(S^{\text{pred}}_i, S^{\text{GT}}).
\end{equation}

\noindent\textbf{Super-parameters.} We optimize our network using the AdamW algorithm with a batch size of 8 and an initial learning rate of $5\times10^{-4}$. The learning rate follows a weight decay of $1\times10^{-4}$ to regularize the optimization process. The model is trained for a total of 50 epochs. During both training and inference, all computations are accelerated on one NVIDIA RTX 5070Ti GPU.

\subsection{Dataset and Evaluation Metrics}
We conducted experiments on four datasets: PKU-LF\cite{gao2023thorough}, DUL-LF\cite{wang2019deep}, HFUT\cite{zhang2017saliency}, and Lytro Illum\cite{zhang2020light}. Among these, PKU-LF contains 6,936 training samples and 2,973 test samples. The model was trained and evaluated on PKU-LF, and further tested on the other three datasets to ensure the reliability of our results.
\begin{figure}[htbp]
\centering
\includegraphics[width=1\columnwidth]{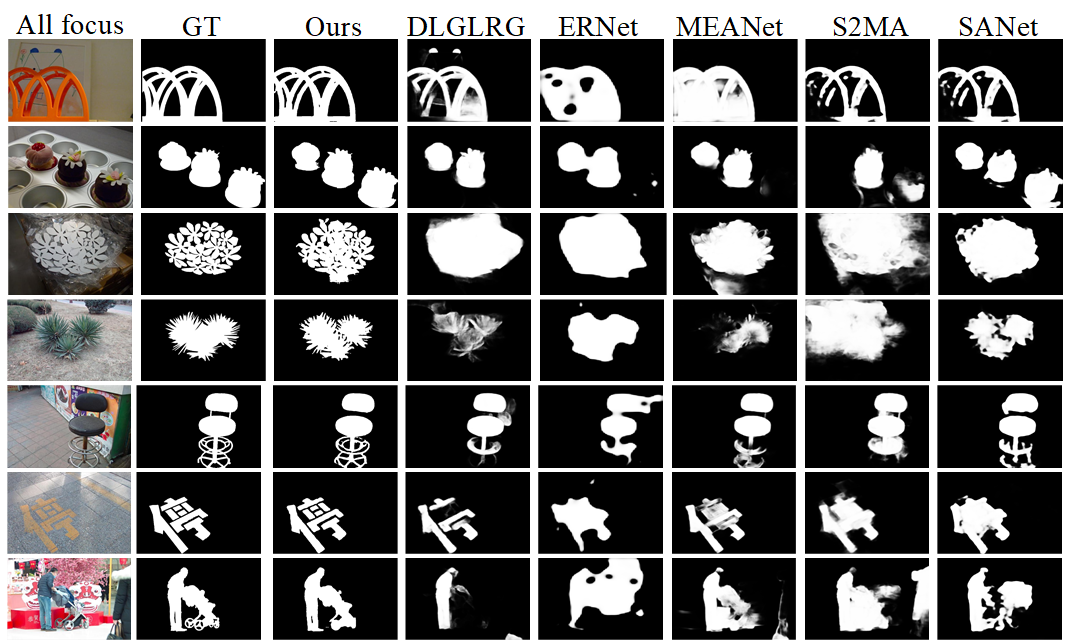}
\caption{Visual comparisons of our methord and SOTA methods.}
\label{fig:4}
\end{figure}

For performance assessment, we adopted the four metrics commonly used in LF SOD, namely S-measure($S_\alpha$), F-measure($F_\beta$), E-measure($E_\phi$), and Mean Absolute Error($M$), to evaluate the model.

\section{EXPERIMENTAL RESULTS}
\label{sec:typestyle}
\subsection{Comparison with State-of-the-Arts}
In comparative analysis, we compared our proposed SPLF-SAM with 12 SOTA LF SOD methods, namely GMERNet\cite{li2024gated}, STSA\cite{gao2023thorough}, MEANet\cite{jiang2022meanet}, JLDCF\cite{fu2021siamese}, UCNet\cite{zhang2021uncertainty}, D3Net\cite{fan2020rethinking}, SANet\cite{zhang2021learning}, DLGLRG\cite{liu2021light}, BBS\cite{fan2020bbs}, S2MA\cite{liu2020learning}, ERNet\cite{piao2020exploit}, and MoLF\cite{wang2019deep}. The comparison was conducted on the four aforementioned datasets using four evaluation metrics: $S_\alpha$, $F_\beta$, $E_\phi$, and $M$. Table \ref{tab:1} presents the comparative results, where the best values are highlighted in red and the second-best in blue. As can be observed, our model achieves nearly the best performance across all datasets. Based on the comparison of the $M$ metric, our model achieved improvements of  \textbf{30.8\%}, \textbf{16.7\%}, \textbf{19.6\%}, and \textbf{26.7\%} over the second-best model across the four datasets, demonstrating exceptional performance.

In qualitative analysis, we conducted a visual comparison between SPLF-SAM and representative methods, as illustrated in Fig. \ref{fig:4}, where the third column displays the saliency maps predicted by our method. As can be observed, our model is capable of not only accurately predicting the structure of salient objects but also precisely delineating their boundaries. In highly challenging scenarios involving multiple objects, complex structures, text, or cluttered backgrounds, our approach demonstrates significant improvements over other methods. For instance, in the case of multi-object recognition (second row), only our model successfully detects all objects completely. Similarly, for complex objects (third and fourth rows), where other models can only identify coarse regions, our method accurately captures object boundaries, demonstrating exceptional precision.

\begin{table}[htbp]
\centering
\caption{Ablation Study Results.}
\scalebox{1}{
\begin{tabular}{@{}p{0.32\textwidth}cc@{}}
\toprule
\multicolumn{1}{c}{\textbf{Models}} & \textbf{$F_\beta\uparrow$} & \textbf{$M\downarrow$} \\
\midrule
(a)\makebox[0.32\textwidth]{Full fine-tuning}          & 0.832 & 0.065 \\
(b)\makebox[0.32\textwidth]{SAM + Decoder*}            & 0.907 & 0.034 \\
(c)\makebox[0.32\textwidth]{SAM + MAFA* + Decoder*}     & 0.904 & 0.030 \\
(d)\makebox[0.32\textwidth]{SAM + MAFA + Decoder*}     & 0.912 & 0.029 \\
(e)\makebox[0.32\textwidth]{SAM + MAFA + Decoder}      & 0.940 & 0.021 \\
\midrule
(f)\makebox[0.32\textwidth]{SAM + MAFA + Decoder + UMFEB} & 0.955 & 0.018 \\
\bottomrule
\end{tabular}}
\\[0.1cm]
\begin{minipage}{0.45\textwidth} 
\raggedright 
\small
MAFA* denotes without fourier filtering.\\ 
Decoder* denotes without receiving prompt information.
\end{minipage}
\label{tab:2}
\end{table}

\subsection{Ablation Study}
To demonstrate the effectiveness of different modules in our model, we conducted ablation experiments, as summarized in Table \ref{tab:2}. The table compares six configurations from top to bottom: full-parameter training, encoder and decoder only, adding the adapter without filtering, adding the adapter with filtering, incorporating the prompt module, and the complete model with all modules. It can be observed that the performance metrics $F_\beta$ and $M$ progressively improve with the incremental addition of each module, confirming the contribution of our proposed components.

\section{Conclusion}
\label{sec:typestyle}
In this paper, we propose a light field salient object detection model called SPLF-SAM, characterized by the MAFA process that separates and enhances high- and low-frequency features from noise. The key component in SPLF-SAM, namely UMFEB, is validated by ablation experiments. The experimental results show that both UMFEB, and self-prompting process, are both essential for obtaining higher detection accuracy.  Our model achieves impressive results compared with multiple SOTA results. In the future, we will attempt to incorporate more light field information to achieve better segmentation performance.

\clearpage
\newpage
\bibliographystyle{IEEEbib}
\bibliography{refs}

\begin{thebibliography}{10}

\bibitem{10446685}
Junyi Wang, Bin Chen, Wenrui Fan, and Yongjiang Liu,
\newblock ``Co-salient object detection via discriminative prototypes contrast,''
\newblock in {\em ICASSP 2024 - 2024 IEEE International Conference on Acoustics, Speech and Signal Processing (ICASSP)}, April 2024, pp. 7390--7394.

\bibitem{JIANG2025117}
Wenhui Jiang, Qi~Shu, Hongwei Cheng, Yuming Fang, Yifan Zuo, and Xiaowei Zhao,
\newblock ``Learning stage-wise fusion transformer for light field saliency detection,''
\newblock {\em Pattern Recognition Letters}, vol. 197, pp. 117--123, 2025.

\bibitem{zhu2024visionmambaefficientvisual}
Lianghui Zhu, Bencheng Liao, Qian Zhang, Xinlong Wang, Wenyu Liu, and Xinggang Wang,
\newblock ``Vision mamba: Efficient visual representation learning with bidirectional state space model,'' 2024.

\bibitem{huang2024saliency}
Rui Huang, Qingyi Zhao, Yan Xing, Sihua Gao, Weifeng Xu, Yuxiang Zhang, and Wei Fan,
\newblock ``A saliency enhanced feature fusion based multiscale rgb-d salient object detection network,''
\newblock in {\em ICASSP 2024-2024 IEEE International Conference on Acoustics, Speech and Signal Processing (ICASSP)}. IEEE, 2024, pp. 9356--9360.

\bibitem{gao2024multiscaledetailenhancedsegmentmodel}
Shixuan Gao, Pingping Zhang, Tianyu Yan, and Huchuan Lu,
\newblock ``Multi-scale and detail-enhanced segment anything model for salient object detection,'' 2024.

\bibitem{rs17020342}
Xiaoning Zhang, Yi~Yu, Daqun Li, and Yuqing Wang,
\newblock ``Progressive self-prompting segment anything model for salient object detection in optical remote sensing images,''
\newblock {\em Remote Sensing}, vol. 17, no. 2, 2025.

\bibitem{gao2023thorough}
Wei Gao, Songlin Fan, Ge~Li, and Weisi Lin,
\newblock ``A thorough benchmark and a new model for light field saliency detection,''
\newblock {\em IEEE Transactions on Pattern Analysis and Machine Intelligence}, 2023.

\bibitem{wang2019deep}
Tiantian Wang, Yongri Piao, Xiao Li, Lihe Zhang, and Huchuan Lu,
\newblock ``Deep learning for light field saliency detection,''
\newblock in {\em Proceedings of the IEEE/CVF International Conference on Computer Vision}, 2019, pp. 8838--8848.

\bibitem{zhang2017saliency}
Jun Zhang, Meng Wang, Liang Lin, Xun Yang, Jun Gao, and Yong Rui,
\newblock ``Saliency detection on light field: A multi-cue approach,''
\newblock {\em ACM Transactions on Multimedia Computing, Communications, and Applications (TOMM)}, vol. 13, no. 3, pp. 1--22, 2017.

\bibitem{zhang2020light}
Jun Zhang, Yamei Liu, Shengping Zhang, Ronald Poppe, and Meng Wang,
\newblock ``Light field saliency detection with deep convolutional networks,''
\newblock {\em IEEE Transactions on Image Processing}, vol. 29, pp. 4421--4434, 2020.

\bibitem{li2024gated}
Yefan Li, Fuqing Duan, and Ke~Lu,
\newblock ``Gated multi-modal edge refinement network for light field salient object detection,''
\newblock {\em ACM Transactions on Multimedia Computing, Communications and Applications}, vol. 20, no. 10, pp. 1--20, 2024.

\bibitem{jiang2022meanet}
Yao Jiang, Wenbo Zhang, Keren Fu, and Qijun Zhao,
\newblock ``Meanet: Multi-modal edge-aware network for light field salient object detection,''
\newblock {\em Neurocomputing}, vol. 491, pp. 78--90, 2022.

\bibitem{fu2021siamese}
Keren Fu, Deng-Ping Fan, Ge-Peng Ji, Qijun Zhao, Jianbing Shen, and Ce~Zhu,
\newblock ``Siamese network for rgb-d salient object detection and beyond,''
\newblock {\em IEEE transactions on pattern analysis and machine intelligence}, vol. 44, no. 9, pp. 5541--5559, 2021.

\bibitem{zhang2021uncertainty}
Jing Zhang, Deng-Ping Fan, Yuchao Dai, Saeed Anwar, Fatemeh Saleh, Sadegh Aliakbarian, and Nick Barnes,
\newblock ``Uncertainty inspired rgb-d saliency detection,''
\newblock {\em IEEE transactions on pattern analysis and machine intelligence}, vol. 44, no. 9, pp. 5761--5779, 2021.

\bibitem{fan2020rethinking}
Deng-Ping Fan, Zheng Lin, Zhao Zhang, Menglong Zhu, and Ming-Ming Cheng,
\newblock ``Rethinking rgb-d salient object detection: Models, data sets, and large-scale benchmarks,''
\newblock {\em IEEE Transactions on neural networks and learning systems}, vol. 32, no. 5, pp. 2075--2089, 2020.

\bibitem{zhang2021learning}
Yi~Zhang, Geng Chen, Qian Chen, Yujia Sun, Yong Xia, Olivier Deforges, Wassim Hamidouche, and Lu~Zhang,
\newblock ``Learning synergistic attention for light field salient object detection,''
\newblock {\em arXiv preprint arXiv:2104.13916}, 2021.

\bibitem{liu2021light}
Nian Liu, Wangbo Zhao, Dingwen Zhang, Junwei Han, and Ling Shao,
\newblock ``Light field saliency detection with dual local graph learning and reciprocative guidance,''
\newblock in {\em Proceedings of the IEEE/CVF international conference on computer vision}, 2021, pp. 4712--4721.

\bibitem{fan2020bbs}
Deng-Ping Fan, Yingjie Zhai, Ali Borji, Jufeng Yang, and Ling Shao,
\newblock ``Bbs-net: Rgb-d salient object detection with a bifurcated backbone strategy network,''
\newblock in {\em European conference on computer vision}. Springer, 2020, pp. 275--292.

\bibitem{liu2020learning}
Nian Liu, Ni~Zhang, and Junwei Han,
\newblock ``Learning selective self-mutual attention for rgb-d saliency detection,''
\newblock in {\em Proceedings of the IEEE/CVF conference on computer vision and pattern recognition}, 2020, pp. 13756--13765.

\bibitem{piao2020exploit}
Yongri Piao, Zhengkun Rong, Miao Zhang, and Huchuan Lu,
\newblock ``Exploit and replace: An asymmetrical two-stream architecture for versatile light field saliency detection,''
\newblock in {\em Proceedings of the AAAI Conference on Artificial Intelligence}, 2020, vol.~34, pp. 11865--11873.

\end{thebibliography}

\end{document}